\documentclass{article} 
\usepackage{iclr2025_conference,times}

\usepackage{amsmath,amsfonts,bm}

\def\eqref#1{equation~\ref{#1}}

\def\1{\bm{1}}

\DeclareMathAlphabet{\mathsfit}{\encodingdefault}{\sfdefault}{m}{sl}
\SetMathAlphabet{\mathsfit}{bold}{\encodingdefault}{\sfdefault}{bx}{n}

\usepackage{hyperref}
\usepackage{url}

\usepackage{graphicx}

\usepackage{booktabs}

\usepackage{mdframed}

\usepackage{listings}

\usepackage{enumitem}

\newmdenv[
  backgroundcolor=gray!10,  
  linecolor=black,          
  linewidth=0.5pt,          
  roundcorner=4pt,          
  skipabove=10pt,           
  skipbelow=10pt,           
  innertopmargin=10pt,      
  innerbottommargin=10pt,   
  innerleftmargin=10pt,     
  innerrightmargin=10pt,    
  splitbottomskip=10pt,     
  splittopskip=10pt         
]{promptbox}

\usepackage{amsmath,amssymb} 

\newcommand{\indic}[1]{\left[ #1 \right]}

\usepackage{multirow}

\usepackage{hyperref}
\usepackage{url}

\title{Towards Lighter and Robust Evaluation for Retrieval Augmented Generation}

\iclrfinalcopy

\author{
Alex-Răzvan Ispas \\
Paris-Saclay University, BNP Paribas CIB\\
\texttt{alex.razvan.ispas@gmail.com} \\
\And
Charles-Élie Simon  \\
BNP Paribas CIB \\
\texttt{charleselie.simon@bnpparibas.com} \\
\And
Fabien Caspani \\
BNP Paribas CIB \\
\texttt{fabien.caspani@bnpparibas.com} \\
\And
Vincent Guigue \\
AgroParisTech \\
\texttt{vincent.guigue@agroparistech.fr} \\
}

\begin{document}

\maketitle

\begin{abstract}
Large Language Models are prompting us to view more NLP tasks from a generative perspective. At the same time, they offer a new way of accessing information, mainly through the RAG framework. While there have been notable improvements for the autoregressive models, overcoming hallucination in the generated answers remains a continuous problem. A standard solution is to use commercial LLMs, such as GPT4, to evaluate these algorithms. However, such frameworks are expensive and not very transparent. Therefore, we propose a study which demonstrates the interest of open-weight models for evaluating RAG hallucination.
We develop a lightweight approach using smaller, quantized LLMs to provide an accessible and interpretable metric that gives continuous scores for the generated answer with respect to their correctness and faithfulness.
This score allows us to question decisions' reliability and explore thresholds to develop a new AUC metric as an alternative to correlation with human judgment. We also provide a script for the corrected dataset which is available at: \href{https://github.com/Razvanip13/Towards_Lighter_And_Robust_Evaluation}{GitHub Repository}
\end{abstract}

\section{Introduction}

\par Large Language Models (LLMs) have advanced the field of Natural Language Processing (NLP) in recent years  \cite{achiam2023gpt,touvron2023llama,jiang2024mixtral}.
However, some questions require information outside the knowledge scope of the model. 
Therefore, Retrieval Augmented Generation (RAG) \cite{lewis2020retrieval} was proposed to enhance the quality of the answers for questions by retrieving information from a relevant knowledge base.  RAG reliability remains a critical concern, particularly due to hallucinations in the generated answers. While much effort has been dedicated to improving model accuracy, a structured evaluation framework that explicitly addresses hallucination detection is still needed. 

\par In general, we want to assess the quality of an LLM answer by comparing it to a ground truth. Previous approaches used token-based metrics such as Exact Matching (EM) \cite{rajpurkar2016squad}, which are fast to compute but not robust to semantic variations. Human evaluators are considered the most consistent approach for open evaluation, but they require a lot of resources \cite{islam2023financebench}. 
Information extraction approaches can be used to compare the facts in the generated response with a ground truth \cite{jayanthi2021evaluating}, but such systems are quite domain-sensitive, and the metric may not be very robust to domain changes. NLI approaches \cite{liu2021natural} can be used to validate statements, but typically in the context of simple sentences or using a combinatorial approach.
LLM evaluators have recently risen in popularity, and they have been compared with human evaluators regarding agreement on different benchmarks such as MT-bench \cite{zheng2024judging}. Factual few-shot prompting evaluation methods \cite{es2023ragas,zhang2023interpretable,manakul2023selfcheckgpt} can check the level of truth of each answer statement while avoiding the search complexity of a knowledge graph \cite{sansford2024grapheval}.

\par While factual evaluation exhibits progress, most proposed methods rely on enterprise models such as GPT4 \cite{es2023ragas,manakul2023selfcheckgpt}, making the reproduction of experiments challenging due to financial and ethical considerations, such as the protection of sensitive data. Since numerous open-source lightweight LLMs offer comparable performance and are more accessible with quantization, we explore their capabilities with respect
to enterprise models in terms of human agreement.
In the meantime, we look for interpretability by providing an analysis at the statement level for each answer.

Our contributions are as follows:
\begin{itemize}[left=0pt]
    \setlength{\itemsep}{1pt}
    \setlength{\parskip}{1pt}
    \item Proposing a reproducible framework for evaluating RAG-generated answers based on quantized models.
    \item Defining a transparent and interpretable metric based on the decomposition and annotation of the response in terms of statements.
    \item Providing the community with a corrected dataset for RAG evaluation.
    \item Analyze the behaviour of the continuous metric to build a new alignment score with human judgment based on AUC.
\end{itemize}

\section{Related Work} 
\label{sec:related_work}

\par Previous research on Question-Answering (QA), such as SQuAD \cite{rajpurkar2016squad}, relied on deterministic approaches like exact matching (EM) and F1-score. The main advantages of deterministic metrics are that they guarantee correctness, are interpretable, and can be easily generalised. Although traditional metrics like BLEU \cite{papineni2002bleu} and ROUGE \cite{lin2004rouge} perform well for specific tasks such as translation and summarization, they are not robust to task variations. Additionally, metrics that rely on token overlap cannot capture the semantic similarities between two answers. The standard method of evaluating text quality is through human evaluation. Human annotators have the advantage of understanding the context and human preferences much better. However, this method can introduce human bias, and the annotation process requires a significant amount of time and budget.

\par Large Language Models used as evaluators are designed to act as a bridge between the previous approaches. They are capable of handling a wider range of open-ended NLP tasks that are in line with human preferences and that are more difficult to evaluate using traditional metrics \cite{zheng2024judging}. The analysis conducted by \cite{chiang2023can} highlights that LLMs can consistently assess text quality, like humans, when provided with the same instructions and examples. Unlike human evaluation, LLMs used as judges are a scalable alternative that minimizes the need for human involvement during the evaluation process and provides faster model iterations. However, positional bias makes the comparison of the generated answers challenging \cite{wang2023large}. According to \cite{zheng2024judging}, there are three different ways of structuring the prompts. In \emph{pairwise comparison}, the model receives a question and two answers and is asked to choose the better one. For \emph{pointwise scoring}, the model is given a single answer and is asked to assign a score \cite{manakul2023selfcheckgpt}. \emph{Reference-guided scoring} provides the judge with a reference solution in addition to the question and the answer  \cite{es2023ragas}. All of them can enhance their reasoning precision by applying chain-of-thought \cite{wei2022chain}.

\par In the case of RAG evaluation, \cite{es2023ragas} tried to overcome positional bias \cite{wang2023large} by using factual evaluation \cite{manakul2023selfcheckgpt}, increasing the faithfulness accuracy of GPT3.5 from 0.72\% to 0.95\%. Other research \cite{adlakha2023evaluating} compared the deterministic metrics and the LLM evaluators by computing the correlation with the human evaluators. They emphasised that bag-of-tokens recall is highly correlated with humans when the verbose information of the answer is ignored during the evaluation. At the same time, contextual embedding transformers such as BERTScore (BertS) have a lower correlation when they are used to calculate the recall \cite{zhang2019bertscore}. Overall, GPT4 remains one of the highest-rated evaluators in most of the previous studies \cite{zheng2024judging}. 


\section{Methodology}
\label{sec:methodlogy}

\begin{figure}[t]  
  \centering
  \includegraphics[width=\textwidth]{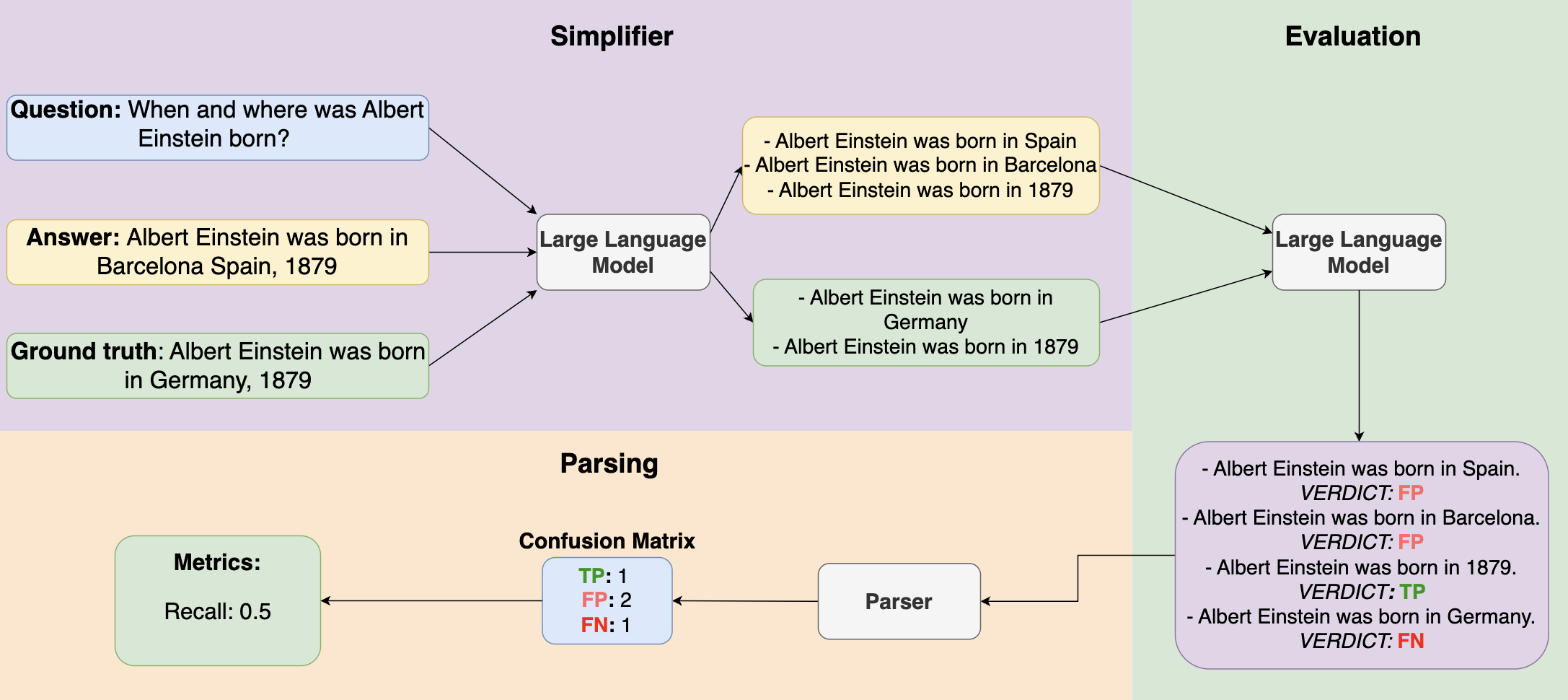}  
  \caption{Evaluation pipeline for answer correctness. First, the simplifier extracts the statements of the answer and the ones of the ground truth. Afterwards, the evaluator labels the statements according to the definitions. 
  Finally, the parser extracts the labelled statements and calculates the metric.}
  \label{fig:evaluation_pipeline}
\end{figure}

\subsection{Factual Evaluation}

\par Following the evaluation methods emphasized by \cite{manakul2023selfcheckgpt} and \cite{es2023ragas}, we will evaluate the answers using metrics that can be grounded in \emph{facts}. We define a \emph{fact} or a \emph{statement} as a declarative sentence that conveys information which can be either true or false. For the example shown in Figure \ref{fig:evaluation_pipeline}, it can be noticed that the answer \emph{Albert Einstein was born in Barcelona Spain, 1879} was split into three statements: \emph{Albert Einstein was born in Spain}, \emph{Albert Einstein was born in Barcelona}, \emph{Albert Einstein was born in 1879}. The label of each statement will be assigned with respect to either a ground truth or a context passage, depending on the calculated metric. By breaking the generated answers into smaller units, we can reduce the complexity of evaluating the truthfulness of each individual fact.

\subsubsection{Correctness}

The first metric is the \emph{answer correctness}. In a RAG setting, an answer is considered \emph{correct} if the statements from the ground truth directly support the statements of the answer. Section \ref{appendix:confusion_matrix_correctness} provides the definitions for the labels which can be assigned to the statements. The final score can be calculated as either the recall or the f1 score of the classified statements.  The recall is a softer version of correctness because it ignores false positive statements, which are verbose information that appears in the answer but not in the ground truth. For instance, if an LLM needs to answer the question \emph{In which country was Albert Einstein born?}, a possible answer is \emph{Germany}. However, an LLM would provide more information, such as Germany's population, which is irrelevant to the ground truth answer. The \emph{harsher} version is the f1-score, which penalises answers that contain verbose information. In general, the f1-score is more suitable for domain adaptation scenarios as the answer should match the structure of the ground truth. Meanwhile, the recall is ideal for scenarios where we seek specific information without being concerned about additional details. Since we work in a few-shot setting, and the chosen dataset was not annotated without considering the verbose information, we will use the recall and ignore the additional information. 

\vspace{-1mm}

\begin{equation*}
\begin{aligned}
    \text{Recall} &= \frac{TP}{TP + FN} \quad & \text{F1} &= \frac{TP}{TP + 0.5 \cdot (FP + FN)}
\end{aligned}
\end{equation*}

\subsubsection{Faithfulness}

\par Instead of quantifying the hallucination of an answer, we are going to quantify its inverse: the \emph{faithfulness} score. We consider that an answer is \emph{faithful} if its claims can be directly inferred in the retrieved context. For the example in Figure \ref{fig:faithfulness_pipeline}, the context passage that mentions information about Einstein's childhood in Germany makes the statements that mention Barcelona and Spain not faithful. The final score can be calculated as the precision of the \emph{passed} and \emph{failed} statements. Section \ref{appendix:confusion_matrix_faithfulness} provides the definitions of the two labels. 

\vspace{-1mm}

\begin{align*}
    Precision = \frac{Passed}{Passed + Failed}
\end{align*}

\subsection{Formalizing}

Our LLM evaluation framework consists of three components. The first one is the  \emph{simplifier} $S$, which transforms a text into a set of elementary statements.
The second one is the \emph{evaluator} $E$, which assesses the matching between the system's response and the ground truth. Both the \emph{simplifier} and the \emph{evaluator} phases are performed by an LLM.
The third one is the \emph{parser} $P$, which extracts the classified statements to calculate the final score. Figure \ref{fig:evaluation_pipeline} showcases the steps taken by the evaluation pipeline to calculate the answer's correctness.

Given a  RAG answer $t$ and ground truth $t'$:
\begin{enumerate}
    \item[$S$]: Transform $t$ and $t'$ into lists of facts as in \cite{manakul2023selfcheckgpt,es2023ragas}, which are expressed in the form of elementary sentences that we call \textit{statements}. $S: t \mapsto \{s_i\}_{i=1}^N$
    
    \item[$E$]: Use few-shot examples to compute semantically which facts are supported or not between the two texts. 
    By default, $s$ is binary since statements can be supported in both directions. However, we are switching to a 3-label system because non-support is not interpreted the same way in both directions: (TP: $s_i \in \{s'_j\}_{j=1}^{N'} $), (FP: $s_i \notin \{s'_j\}_{j=1}^{N'}$), (FN: $s'_j \notin \{s_i\}_{i=1}^{N}$).
    \item[$P$]: As the $E$ step is performed by an LLM, the TP, FP, FN labels are in text format and must be extracted before we can calculate the final metric. 
\end{enumerate}
The final score of \textbf{correctness} is the recall of the labelled statements. This score will be compared with the binary human judgment.

In the case of \textbf{faithfulness}, given a context $c$ and one faithful and unfaithful answer, $t_g$ and $t_p$, the task is to distinguish the most faithful answer.
\begin{enumerate}
    \item[$S$]: Simplify  $t_g$ and $t_p$ to get their statements $\{s_{g,i}\}_{i=1}^N$ $\{s_{p,j}\}_{j=1}^M$
    \item[$E$]: Evaluate the statements individually with respect to the given context $(\{s_{g,i}\}_{i=1}^N, c)$ and $(\{s_{p,i}\}_{i=1}^M, c)$ to get binary labels :(TP: $s_i \Leftarrow c $), (FP: $s_i \not\Leftarrow c$)
    \item[$P$]: Parse the final labels and compute the precision scores to rank the two answers.
    This setting may lead to \emph{ties} which are discussed in Section \ref{ch:wiki_eval}
\end{enumerate}

\par We are mainly focused on cases where the golden passage was provided to generate the answer. Like this, any inaccuracy or hallucination in the generated answer can be attributed to the LLM rather than the retriever. The prompts used in each phase can be consulted in Section \ref{appendix:prompts} of the Appendix. 

\subsection{Parser}

\par The chain-of-thought can increase the reasoning abilities of the LLMs \cite{wei2022chain}. However, there can still be variations in the final results, especially when working with quantized LLMs that have lower precision. Therefore, instead of letting the LLM evaluator calculate the final metric, we parse the labelled statements and manually compute the metrics. 

\par We propose two options to parse the output. The first option is the deterministic approach, which tries to match the labelled statements through a regular expression. This method is fast to compute but can create errors in the final result if the generated answers have variations. The second approach is the constrained generation, which forces the LLM to generate an output that respects a JSON schema. Given a JSON structure, a finite automata is created. During the generation process of each token, a mask will be applied on each invalid token during the respective step in the automata. The process continues until the automata reaches an end state which represents a valid generation that respects the structure. Once the statements are extracted, the confusion matrix can be calculated deterministically. This method implies an additional generation. Consequently, it can take more time to calculate the final metric. At the same time, it is non-deterministic due to its sampling nature. In this work, we will perform constrained generation with the help of the Outlines library \cite{willard2023efficient}. Figure \ref{fig:parsing_strategy} offers an example of the two approaches. The JSON schemas for \emph{correctness} and \emph{faithfulness} are presented in Section \ref{appendix:json}.

\section{Datasets}
\label{sec:datasets}

\subsection{Natural Questions}
\par For evaluating answer correctness, we used InstructQA \cite{adlakha2023evaluating}, which contains questions extracted from three QA datasets: Natural Questions \cite{kwiatkowski2019natural}, HotpotQA \cite{yang2018hotpotqa} and TopioCQA \cite{adlakha2022topiocqa}. More precisely, we used the subset from Natural Questions, which included 100 questions answered by four different LLMs, generating 400 samples in the oracle setting of RAG. Subsequently, a group of students annotated the LLM answers as either correct or incorrect, reaching an inter-annotator agreement of 92.42\%. As the provided data contained mismatching labels, we carefully reviewed all 400 samples and rectified the discrepancies in the annotations. Some of the most common errors included labelling questions as correct, even when the answer and the ground truth contained completely different information. Furthermore, we excluded one question due to its unclear formulation, rendering it unanswerable even for humans. The final reannotated dataset comprised 396 samples.

\par In this setting, given the question, the generated answer and the ground truth, the evaluator needs to identify if the answer is correct or incorrect. This dataset was used to calculate the correlation with respect to the human annotators. Thus, we will use the Spearman and Kendall correlation. Unlike Pearson correlation, Spearman's $\rho$ and Kendall's $\tau$ are non-parametric rank correlation measures. That means we can assess the relationship between the human and the LLM score by using a monotonic function rather than assuming a linear relationship.

\par The human annotators assign a score of 1 for correct and 0 for incorrect, while the LLM evaluator assigns a score between 0 and 1 depending on the level of truth of each statement. Therefore, we will calculate the average F1 score for different threshold levels (F1 AUC). Unlike the Area Under the Curve of Precision and Recall, which shows the balance between the two metrics, F1 AUC can quantify how well the evaluator can separate the score distributions of the correct and incorrect answers by penalising the outliers at each threshold:

\[
F1_{\text{AUC}} = \frac{1}{10} \sum_{i=0}^{10} F1\left(d\left(\mathbf{r}, \frac{i}{10}\right), \mathbf{h}\right), \qquad
d(r, \text{th}) =
\begin{cases} 
    1, & \text{if } r \geq \text{th}, \\
    0, & \text{if } r < \text{th}.
\end{cases}
\]

\vspace{-2mm}
\text{where $r$ is the correctness score of the pipeline, $h$ is the ground truth and $th$ is the current threshold.}
%

\subsection{WikiEval}
\label{ch:wiki_eval}

\par For evaluating faithfulness, we will use the dataset proposed by \cite{es2023ragas}. The dataset contains questions formulated from 50 Wikipedia pages created in 2022. GPT3.5 was used to answer the questions once given context and once not given any context. Afterwards, two human annotators had to decide which one of the two answers was more faithful, reaching a human agreement of 95\%. 

\par In the setting of WikiEval, given a question, the two answers and the context, the evaluator needs to identify which one of the two answers is more faithful. Since we want to compare with the human evaluators, the accuracy score is the number of times the LLM evaluator succeeded in identifying the faithful response divided by the total number of questions. That could also be quantified as the total number of times the good answer's faithfulness score was greater than the faithfulness score of the poor answer.

\par One issue that can be encountered are the \emph{ties}. A \emph{tie} is encountered when the faithfulness scores of the \emph{good answer} and the \emph{poor answer} are \textbf{the same}. Therefore, we propose a score system to handle this problem. Whenever the faithfulness score of the good answer is greater than the faithfulness score of the poor answer, it receives 1 point. If the scores are equal, we assign a partial point of 0.5. Otherwise, it receives 0 points. We provide scores for three scenarios. The worst case is calculated by assigning 1 point only if the faithfulness score of the good answer is strictly greater than the poor ones. The middle case of faithfulness is the sum of all points divided by the number of questions.  The best case assigns 1 point even for the \emph{tie} scenarios:

\[
\begin{aligned}
&s(a_1, a_2) =
\begin{cases} 
    1 & \text{if } faith(a_1) > faith(a_2) \\
    0.5 & \text{if } faith(a_1) = faith(a_2) \\
    0 & \text{if } faith(a_1) < faith(a_2)
\end{cases}
\end{aligned}
\quad
\begin{aligned}
worst_{faith} &= \frac{\sum_{i=1}^n 1_{\indic{faith(good_i) > faith(poor_i)}}}{n} \\
middle_{faith} &= \frac{\sum_{i=1}^n s(good_i, poor_i)}{n} \\
best_{faith} &= \frac{\sum_{i=1}^n 1_{\indic{faith(good_i) \ge faith(poor_i)}}}{n}
\end{aligned}
\]

\section{Experiments}
\label{sec:experiments}


\par The experiments were conducted only on quantized LLMs. For 4-bit precision, we used V100, while the 16 float precision LLMs were running on an A100. The temperature was set to 1 for each LLM evaluator to favour sampling. For each phase of the evaluation pipeline, the LLMs received few-shot examples to learn how to split the answers into statements and how to give a verdict. We implied only two families of LLMs: Llama and Gemma. For the parsing procedure, we used two regular expressions and one constrained generation procedure presented in \ref{sec:appendix_parsing}. Since the constrained generation implies copying each token of a statement in a list, that would require too much time during the sampling process. Therefore, we map each statement to a single token to reduce the sampling process, as shown in Section \ref{sec:appendix_constrained}. An experiment lasted between 4 and 8 hours, depending on the chosen parsing strategy and the size of the model. model. Each LLM evaluator was compared with at least one deterministic metric on human agreement. We consider an evaluator to be robust if its judgement aligns with that of human annotators through correlation and if it has a high separability between the scoring distribution of the good and bad answers through the F1-AUC for correlation and the score system defined in Section \ref{ch:wiki_eval} for faithfulness.

\paragraph{Results Correctness}

\par Table \ref{tab:results} displays the experiments conducted for answer correctness on the Natural Questions subsample of InstructQA. The baseline is the bag-of-tokens recall, which we recalculated to emphasise that the correlation was not altered drastically compared to the original dataset results declared by \cite{adlakha2023evaluating}. In the original paper, the Natural Question subsample has a Spearman $\rho$ of 55.02\%, while our reannotated data has a correlation score of 56.89\% for the same subsample. Since the correlation with the human annotators does not emphasise how well an evaluator can separate the distribution of correct and incorrect answers, we calculate the F1 AUC to estimate how well the evaluator performs at different threshold levels. We prefer the F1 score over the Precision-Recall AUC because it captures the capability of the evaluator to minimize the False Positives and False Negatives rather than the ability to balance the tradeoff between precision and recall. The best configurations have the mean and the standard deviation calculated on five distinct reruns.

\begin{table*}[t]
\caption{Correctness and faithfulness experiments. The evaluator represents either a deterministic metric or an LLM. BoT stands for bag-of-token. L3 stands for Llama3 and G2 for Gemma2. The Parsing column denotes the parsing strategy: R1 for the first regular expression, R2 for the second one and C for constrained generation. F1 AUC is the area under the curve of the F1-score. $\rho$ and $\tau$ are the Spearman and Kendall correlation. The last columns are the lower bound, the mean and the upper bound for the Wikieval questions. The best configurations are in bold}

\begin{center}
\begin{small}
\begin{sc}
\resizebox{.99\textwidth}{!}{
\begin{tabular}{cc|ccc|cccc}

\toprule
 \multicolumn{2}{c|}{Pipeline} &\multicolumn{3}{|c|}{Correctness} & \multicolumn{3}{|c}{Faithfulness} \\
\midrule
 Evaluator  & Parsing & F1 AUC & $\rho$ &  $\tau$ & Worst& Middle & Best \\
\midrule

BoT Recall  & N/A & 88.78 & 56.89 & 52.89 & N/A& N/A& N/A \\
RAGAS (GPT3.5-Turbo) & N/A & N/A & N/A & N/A & N/A & N/A &0.95 \\
K-precision  & N/A & N/A & N/A & N/A & N/A& N/A&0.96 \\

L3 8B 4 bit & R1& 87.44 & 30.21 & 28.54& 0.74  & 0.84 & 0.94&  \\
L3 8B 4 bit & R2 & 89.62 & 37.36 & 36.54 & 0.78 & 0.85 & 0.92 \\
L3 8B 4 bit & C & 90.57 & 44.41 & 43.28 &0.72 & 0.89 & 1.0\\
L3.1 8B 4 bit & R1 &86.14 &  36.89 & 34.34 & 0.74 & 0.79  & 0.84   \\
L3.1 8B 4 bit & R2 & 86.47 & 40.02 & 37.74 &0.78 & 0.82 & 0.86  \\
L3.1 8B 4 bit& C& 75.33 & 30.84 & 29.50 &0.72 & 0.83 & 0.94 \\
G2 9B 4 bit & R1 &  92.20 & 52.55 & 50.21 & \textbf{0.92} & \textbf{0.94} & \textbf{0.96}   \\
G2 9B 4 bit & R2  & \textbf{93.83} $\pm$\textbf{0.27} & \textbf{62.06} $\pm$\textbf{1.83}& \textbf{60.49} $\pm$\textbf{1.54} &\textbf{0.82} & \textbf{0.88} & \textbf{0.94}\\
G2 9B 4 bit & C & 89.05 & 55.01 & 52.10  &0.82 & 0.90 & 0.98   \\
L3 70B 16 bit & R1& 86.42 & 49.44 & 45.41& 0.94  & 0.95 & 0.96 \\
L3 70B 16 bit & R2 & \textbf{92.72} $\pm$\textbf{0.20}  & \textbf{63.59} $\pm$\textbf{1.51} & \textbf{60.55} $\pm$\textbf{1.39} &\textbf{0.94} & \textbf{0.95} & \textbf{0.96}\\
L3 70B 16 bit& C& 77.21 & 40.52 & 37.23 &0.88 & 0.91  & 0.94 \\ 
\bottomrule
\end{tabular}
}
\end{sc}
\end{small}
\end{center}
\vskip -0.1in
\label{tab:results}
\end{table*}

\begin{figure*}[t]  
  \centering
  \includegraphics[width=\textwidth]{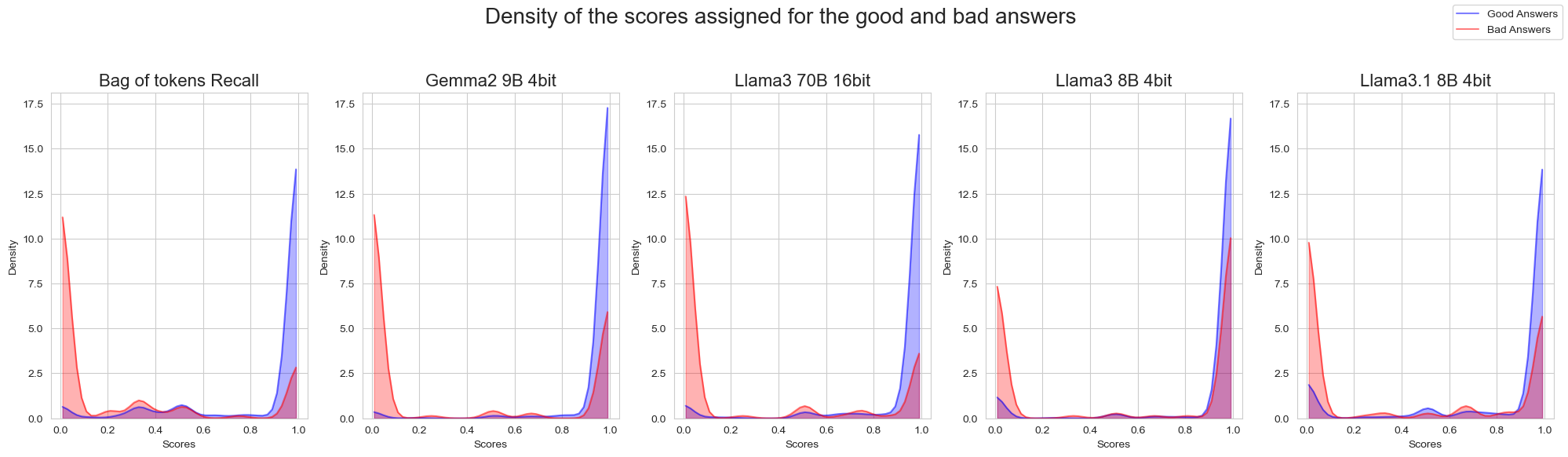}  
  \caption{The density distribution plots of the correctness evaluators that use the second regular expression as parser. The distribution of the correct and incorrect answers are marked with blue and red, respectively. The labels were chosen according to the human annotations.}
  \label{fig:density_correctness}
\end{figure*}

\par For the parsing procedure, it could be noticed that constrained generation is not always the optimal solution. While Llama3 8B benefited from constrained generation parsing, reaching an AUC F1 of 90.57\% and a Spearman correlation of 44.41, the other LLMs had better results using regular expressions. This result emphasises that deterministic parsing can be a faster alternative which does not require guided generated answers.

\par Although the bag-of-tokens recall has a high correlation with human annotators, the AUC F1 is much lower compared to Gemma2 9B and Llama3 70B. In Figure \ref{fig:density_correctness}, we can see the distribution of correct and incorrect answers for the evaluators, which use the second regular expression as the parser, and the distribution of the bag-of-tokens recall. While the bag-of-tokens recall can separate the distribution of the incorrect answers, the scores of the correct answers are more dispersed, resulting in a lower F1 AUC when the threshold is increased. Bag-of-tokens is a lexically based metric, and if the generated answer does not use the exact words as the ones in the ground truth, it cannot capture the semantic similarities, resulting in a penalised score.

\par One impressive result is the Gemma2 9B, which uses the deterministic parser, as it performs similarly to Llama3 70B, although it is about seven times lighter and has four times lower precision. In Figure \ref{fig:density_correctness}, we can observe that Gemma2 is usually very confident in its decisions, assigning a score of either 0 or 1 almost all the time. However, it misclassified a considerable percentage of incorrect answers by giving them a score of 1. Llama3 70B minimizes the number of wrong answers that have a score of 1, but the distribution of the incorrect answers remains dispersed. Nevertheless, both Gemma2 9B and Llama3 70B can maintain the distribution of good answers above the score of 0.5. Unfortunately, Llama3 8B cannot distinguish between correct and incorrect answers, assigning a score of 1 almost every time. Llama3.1 has both distributions sparsed between 0 and 1.

\paragraph{Results Faithfulness}

\par Table \ref{tab:results} displays the results for the faithfulness experiments conducted on WikiEval. Each experiment has three calculated scores according to the worst, middle and best case. Since we had minimal information regarding how the final accuracy score was calculated in the original paper \cite{es2023ragas}, we assumed that the best-performing seed was reported, that being the best case score. We also assume that a good evaluator should have a high accuracy score for the lower bound, and it needs to minimize the difference between the best and worst case. The baseline with whom we compare the LLM evaluators are the results declared by \cite{es2023ragas} and the Knowledge-Precision (K-Precision) metric, which counts how many tokens from the generated answer can be found in the retrieved context.

\begin{figure*}[h]  
  \centering
  \includegraphics[width=\textwidth]{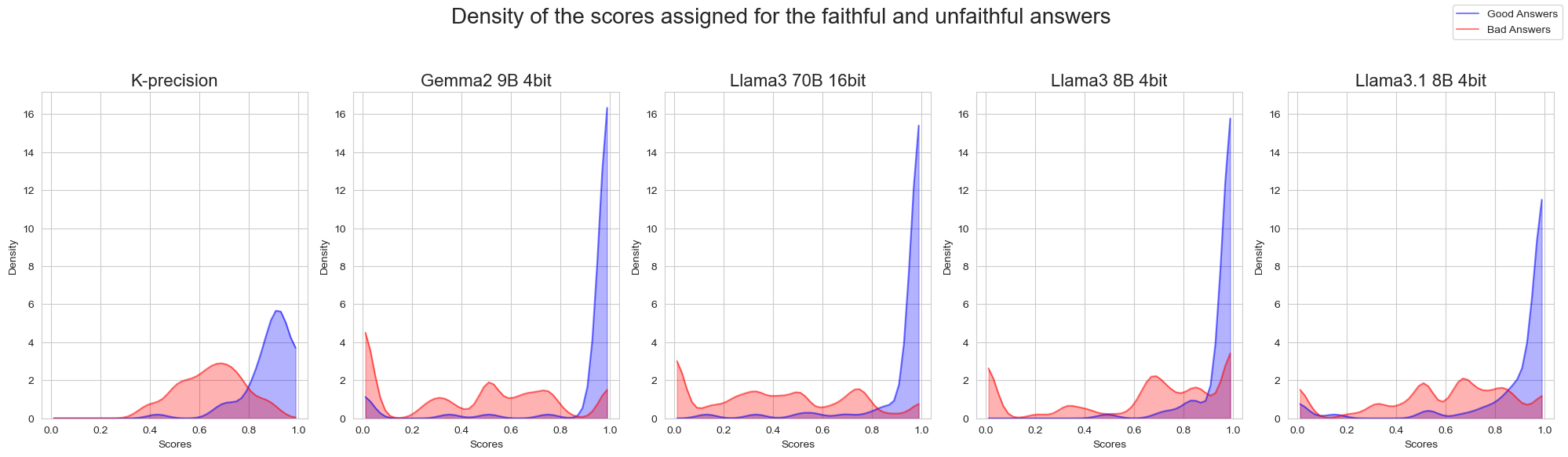}  
  \caption{The density distribution plots of the faithfulness evaluators that use the second regular expression as parsing. The distribution of the faithful and unfaithful answers scores are marked with blue and red, respectively. The labels were chosen according to the human annotations.}
  \label{fig:density_faithfulness}
\end{figure*}

\par If we look at the experiments which use constrained generation, it could be noticed that they have a considerably high score for the worst case. However, the difference between the worst and the best case is the highest among all experiments, which means that it encounters many scenarios where it assigns the same score for both the faithful and unfaithful answers. The deterministic parsers have a much lower difference, regardless of the evaluator. Llama3 70B is the highest-performing evaluator, reaching the lowest possible difference and the highest score for the worst case: 0.94\%. Once again, Gemma2 9B is reaching the second-highest worst case score despite having a lower precision and size than Llama3 70B. 
 
\par By observing the density distributions of the scores assigned to the faithful and unfaithful answers in Figure \ref{fig:density_faithfulness}, we can see that the K-precision succeeds in maintaining the scores of the faithful answers above 0.5. Unfortunately, the distribution of the unfaithful answers is shifted towards the middle of the scale, making it difficult to estimate if the answer is wrong just by looking at the score. An answer can contain similar tokens to those existing in the context. However, the semantics of the phrases can be different. Gemma2 9B and Llama3 70B assign a score of 1 almost all the time for the faithful answers, while the scores for the unfaithful answers are more sparse. That is expected, considering that an unfaithful answer can contain some statements that can be inferred from the context. Nevertheless, few unfaithful answers receive a score of 1, making the two distributions separable. Unfortunately, Llama3 8B and Llama3.1 8B have the unfaithful answers scores dispersed across the whole scale. 

\par If we compare with RAGAS, we notice that even the lower bound and the middle accuracy of our best configuration is close to the score of the RAGAS framework, which uses GPT3.5-Turbo as an evaluator. That reinforces the fact that lighter LLMs could become an alternative for having evaluation performance close to one of the enterprise LLMs. K-precision has high accuracy, but it should be taken into account that, in this setting, the faithful answer was selected by comparing its score with the unfaithful one. Therefore, it is difficult to decide if an answer is either faithful or unfaithful just by looking at the score of the answer due to the overlap of the two distributions.

\section{Conclusion}
\label{sec:conclusion}

\par To conclude, we benchmarked the capabilities of lighter and quantized LLMs to evaluate the correctness and faithfulness of answers generated by the RAG pipelines, and we highlighted that quantized Llama3 70B and Gemma2 9B perform similarly to the enterprise LLM evaluators. We also provided statistical interpretations by observing the distribution of the scores assigned to the good and bad answers and we highlighted that the best-performing configurations have a high separability for the two distributions. In general, deterministic parsing approaches should be favoured whenever possible, while constrained generation should be utilised as a last resort. According to our analysis, wrong answers are more challenging to detect just by looking at their score distribution. For future work, we would like to explore mixtures of LLM evaluators to assess if they can outperform individual evaluators in terms of human agreement and reduce bias

\bibliography{iclr2025_conference}

\begin{thebibliography}{24}
\providecommand{\natexlab}[1]{#1}
\providecommand{\url}[1]{\texttt{#1}}
\expandafter\ifx\csname urlstyle\endcsname\relax
  \providecommand{\doi}[1]{doi: #1}\else
  \providecommand{\doi}{doi: \begingroup \urlstyle{rm}\Url}\fi

\bibitem[Achiam et~al.(2023)Achiam, Adler, Agarwal, Ahmad, Akkaya, Aleman, Almeida, Altenschmidt, Altman, Anadkat, et~al.]{achiam2023gpt}
Josh Achiam, Steven Adler, Sandhini Agarwal, Lama Ahmad, Ilge Akkaya, Florencia~Leoni Aleman, Diogo Almeida, Janko Altenschmidt, Sam Altman, Shyamal Anadkat, et~al.
\newblock Gpt-4 technical report.
\newblock \emph{arXiv preprint arXiv:2303.08774}, 2023.

\bibitem[Adlakha et~al.(2022)Adlakha, Dhuliawala, Suleman, de~Vries, and Reddy]{adlakha2022topiocqa}
Vaibhav Adlakha, Shehzaad Dhuliawala, Kaheer Suleman, Harm de~Vries, and Siva Reddy.
\newblock Topiocqa: Open-domain conversational question answering with topic switching.
\newblock \emph{Transactions of the Association for Computational Linguistics}, 10:\penalty0 468--483, 2022.

\bibitem[Adlakha et~al.(2023)Adlakha, BehnamGhader, Lu, Meade, and Reddy]{adlakha2023evaluating}
Vaibhav Adlakha, Parishad BehnamGhader, Xing~Han Lu, Nicholas Meade, and Siva Reddy.
\newblock Evaluating correctness and faithfulness of instruction-following models for question answering.
\newblock \emph{Transactions of the Association for Computational Linguistics}, 2023.

\bibitem[Chiang \& Lee(2023)Chiang and Lee]{chiang2023can}
Cheng-Han Chiang and Hung-yi Lee.
\newblock Can large language models be an alternative to human evaluations?
\newblock \emph{Proceedings of the 61st Annual Meeting of the Association for Computational Linguistics}, 2023.

\bibitem[Es et~al.(2023)Es, James, Espinosa-Anke, and Schockaert]{es2023ragas}
Shahul Es, Jithin James, Luis Espinosa-Anke, and Steven Schockaert.
\newblock Ragas: Automated evaluation of retrieval augmented generation.
\newblock \emph{Proceedings of the 18th Conference of the European Chapter of the Association for Computational Linguistics: System Demonstrations}, 2023.

\bibitem[Islam et~al.(2023)Islam, Kannappan, Kiela, Qian, Scherrer, and Vidgen]{islam2023financebench}
Pranab Islam, Anand Kannappan, Douwe Kiela, Rebecca Qian, Nino Scherrer, and Bertie Vidgen.
\newblock Financebench: A new benchmark for financial question answering.
\newblock \emph{arXiv preprint arXiv:2311.11944}, 2023.

\bibitem[Jayanthi et~al.(2021)Jayanthi, Embar, and Raghunathan]{jayanthi2021evaluating}
Sai~Muralidhar Jayanthi, Varsha Embar, and Karthik Raghunathan.
\newblock Evaluating pretrained transformer models for entity linking in task-oriented dialog.
\newblock \emph{arXiv preprint arXiv:2112.08327}, 2021.

\bibitem[Jiang et~al.(2024)Jiang, Sablayrolles, Roux, Mensch, Savary, Bamford, Chaplot, Casas, Hanna, Bressand, et~al.]{jiang2024mixtral}
Albert~Q Jiang, Alexandre Sablayrolles, Antoine Roux, Arthur Mensch, Blanche Savary, Chris Bamford, Devendra~Singh Chaplot, Diego de~las Casas, Emma~Bou Hanna, Florian Bressand, et~al.
\newblock Mixtral of experts.
\newblock \emph{arXiv preprint arXiv:2401.04088}, 2024.

\bibitem[Kwiatkowski et~al.(2019)Kwiatkowski, Palomaki, Redfield, Collins, Parikh, Alberti, Epstein, Polosukhin, Devlin, Lee, et~al.]{kwiatkowski2019natural}
Tom Kwiatkowski, Jennimaria Palomaki, Olivia Redfield, Michael Collins, Ankur Parikh, Chris Alberti, Danielle Epstein, Illia Polosukhin, Jacob Devlin, Kenton Lee, et~al.
\newblock Natural questions: a benchmark for question answering research.
\newblock \emph{Transactions of the Association for Computational Linguistics}, 7:\penalty0 453--466, 2019.

\bibitem[Lewis et~al.(2020)Lewis, Perez, Piktus, Petroni, Karpukhin, Goyal, K{\"u}ttler, Lewis, Yih, Rockt{\"a}schel, et~al.]{lewis2020retrieval}
Patrick Lewis, Ethan Perez, Aleksandra Piktus, Fabio Petroni, Vladimir Karpukhin, Naman Goyal, Heinrich K{\"u}ttler, Mike Lewis, Wen-tau Yih, Tim Rockt{\"a}schel, et~al.
\newblock Retrieval-augmented generation for knowledge-intensive nlp tasks.
\newblock \emph{Advances in Neural Information Processing Systems}, 33:\penalty0 9459--9474, 2020.

\bibitem[Lin(2004)]{lin2004rouge}
Chin-Yew Lin.
\newblock Rouge: A package for automatic evaluation of summaries.
\newblock In \emph{Text summarization branches out}, pp.\  74--81, 2004.

\bibitem[Liu et~al.(2021)Liu, Cui, Liu, and Zhang]{liu2021natural}
Hanmeng Liu, Leyang Cui, Jian Liu, and Yue Zhang.
\newblock Natural language inference in context-investigating contextual reasoning over long texts.
\newblock In \emph{Proceedings of the AAAI conference on artificial intelligence}, volume~35, pp.\  13388--13396, 2021.

\bibitem[Manakul et~al.(2023)Manakul, Liusie, and Gales]{manakul2023selfcheckgpt}
Potsawee Manakul, Adian Liusie, and Mark~JF Gales.
\newblock Selfcheckgpt: Zero-resource black-box hallucination detection for generative large language models.
\newblock \emph{Conference on Empirical Methods in Natural Language Processing}, 2023.

\bibitem[Papineni et~al.(2002)Papineni, Roukos, Ward, and Zhu]{papineni2002bleu}
Kishore Papineni, Salim Roukos, Todd Ward, and Wei-Jing Zhu.
\newblock Bleu: a method for automatic evaluation of machine translation.
\newblock In \emph{Proceedings of the 40th annual meeting of the Association for Computational Linguistics}, pp.\  311--318, 2002.

\bibitem[Rajpurkar et~al.(2016)Rajpurkar, Zhang, Lopyrev, and Liang]{rajpurkar2016squad}
Pranav Rajpurkar, Jian Zhang, Konstantin Lopyrev, and Percy Liang.
\newblock Squad: 100,000+ questions for machine comprehension of text.
\newblock \emph{Conference on Empirical Methods in Natural Language Processing}, 2016.

\bibitem[Sansford et~al.(2024)Sansford, Richardson, Maretic, and Saada]{sansford2024grapheval}
Hannah Sansford, Nicholas Richardson, Hermina~Petric Maretic, and Juba~Nait Saada.
\newblock Grapheval: A knowledge-graph based llm hallucination evaluation framework.
\newblock \emph{arXiv preprint arXiv:2407.10793}, 2024.

\bibitem[Touvron et~al.(2023)Touvron, Martin, Stone, Albert, Almahairi, Babaei, Bashlykov, Batra, Bhargava, Bhosale, et~al.]{touvron2023llama}
Hugo Touvron, Louis Martin, Kevin Stone, Peter Albert, Amjad Almahairi, Yasmine Babaei, Nikolay Bashlykov, Soumya Batra, Prajjwal Bhargava, Shruti Bhosale, et~al.
\newblock Llama 2: Open foundation and fine-tuned chat models.
\newblock \emph{arXiv preprint arXiv:2307.09288}, 2023.

\bibitem[Wang et~al.(2023)Wang, Li, Chen, Cai, Zhu, Lin, Cao, Liu, Liu, and Sui]{wang2023large}
Peiyi Wang, Lei Li, Liang Chen, Zefan Cai, Dawei Zhu, Binghuai Lin, Yunbo Cao, Qi~Liu, Tianyu Liu, and Zhifang Sui.
\newblock Large language models are not fair evaluators.
\newblock \emph{arXiv preprint arXiv:2305.17926}, 2023.

\bibitem[Wei et~al.(2022)Wei, Wang, Schuurmans, Bosma, Xia, Chi, Le, Zhou, et~al.]{wei2022chain}
Jason Wei, Xuezhi Wang, Dale Schuurmans, Maarten Bosma, Fei Xia, Ed~Chi, Quoc~V Le, Denny Zhou, et~al.
\newblock Chain-of-thought prompting elicits reasoning in large language models.
\newblock \emph{Advances in neural information processing systems}, 35:\penalty0 24824--24837, 2022.

\bibitem[Willard \& Louf(2023)Willard and Louf]{willard2023efficient}
Brandon~T Willard and R{\'e}mi Louf.
\newblock Efficient guided generation for large language models.
\newblock \emph{arXiv e-prints}, 2023.

\bibitem[Yang et~al.(2018)Yang, Qi, Zhang, Bengio, Cohen, Salakhutdinov, and Manning]{yang2018hotpotqa}
Zhilin Yang, Peng Qi, Saizheng Zhang, Yoshua Bengio, William~W Cohen, Ruslan Salakhutdinov, and Christopher~D Manning.
\newblock Hotpotqa: A dataset for diverse, explainable multi-hop question answering.
\newblock \emph{arXiv preprint arXiv:1809.09600}, 2018.

\bibitem[Zhang et~al.(2023)Zhang, Luo, Chuang, Fang, Gaitskell, Hartvigsen, Wu, Fox, Meng, and Glass]{zhang2023interpretable}
Tianhua Zhang, Hongyin Luo, Yung-Sung Chuang, Wei Fang, Luc Gaitskell, Thomas Hartvigsen, Xixin Wu, Danny Fox, Helen Meng, and James Glass.
\newblock Interpretable unified language checking.
\newblock \emph{arXiv preprint arXiv:2304.03728}, 2023.

\bibitem[Zhang et~al.(2019)Zhang, Kishore, Wu, Weinberger, and Artzi]{zhang2019bertscore}
Tianyi Zhang, Varsha Kishore, Felix Wu, Kilian~Q Weinberger, and Yoav Artzi.
\newblock Bertscore: Evaluating text generation with bert.
\newblock \emph{International Conference on Learning Representations}, 2019.

\bibitem[Zheng et~al.(2024)Zheng, Chiang, Sheng, Zhuang, Wu, Zhuang, Lin, Li, Li, Xing, et~al.]{zheng2024judging}
Lianmin Zheng, Wei-Lin Chiang, Ying Sheng, Siyuan Zhuang, Zhanghao Wu, Yonghao Zhuang, Zi~Lin, Zhuohan Li, Dacheng Li, Eric Xing, et~al.
\newblock Judging llm-as-a-judge with mt-bench and chatbot arena.
\newblock \emph{Advances in Neural Information Processing Systems}, 36, 2024.

\end{thebibliography}
\bibliographystyle{iclr2025_conference}

\appendix

\section{Faithfulness pipeline}

\begin{figure}[h]  
  \centering
  \includegraphics[width=\linewidth]{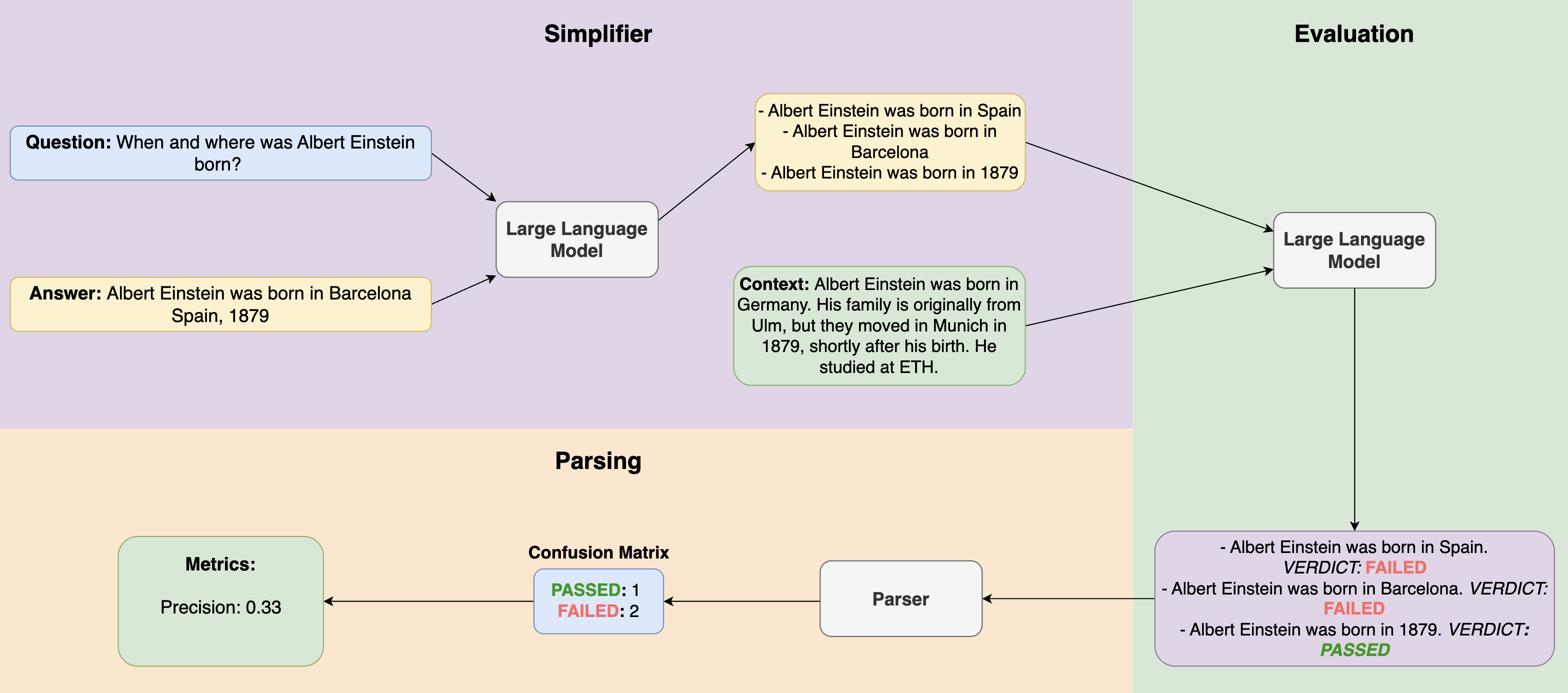}  
  \caption{The evaluation pipeline for faithfulness. Firstly, the LLM extracts the statements of the answer. Afterwards, given the context, the statements are labelled according to the definition from Section \ref{appendix:confusion_matrix_faithfulness}. Finally, the parser extracts the label matches and calculates the metric. }
  \label{fig:faithfulness_pipeline}
\end{figure}

\section{Statements}

\subsection{Statements Correctness}
\label{appendix:confusion_matrix_correctness}

While calculating correctness, a statement can be labelled as:

\begin{itemize}
  \item \textbf{True Positive (TP)}: if the statement appears in the answer and it is directly supported by a statement from the ground truth
  \item \textbf{False Positive (FP)}: if the statement appears in the answer but is not directly supported by a statement from the ground truth
  \item \textbf{False Negative (FN)}: if it appears in the ground truth but does not support any statement from the answer
\end{itemize}

\subsection{Statements Faithfulness}
\label{appendix:confusion_matrix_faithfulness}

While calculating faithfulness, a statement can be labelled as:

\begin{itemize}
  \item \textbf{PASSED}: if the statement can be inferred from the context
  \item \textbf{FAILED}: if the statement cannot be inferred from the context

\end{itemize}

\section{Parsing Strategies}
\label{sec:appendix_parsing}

\subsection{Deterministic Parsing}

\begin{figure*}[t]  
  \centering
  \includegraphics[width=\textwidth]{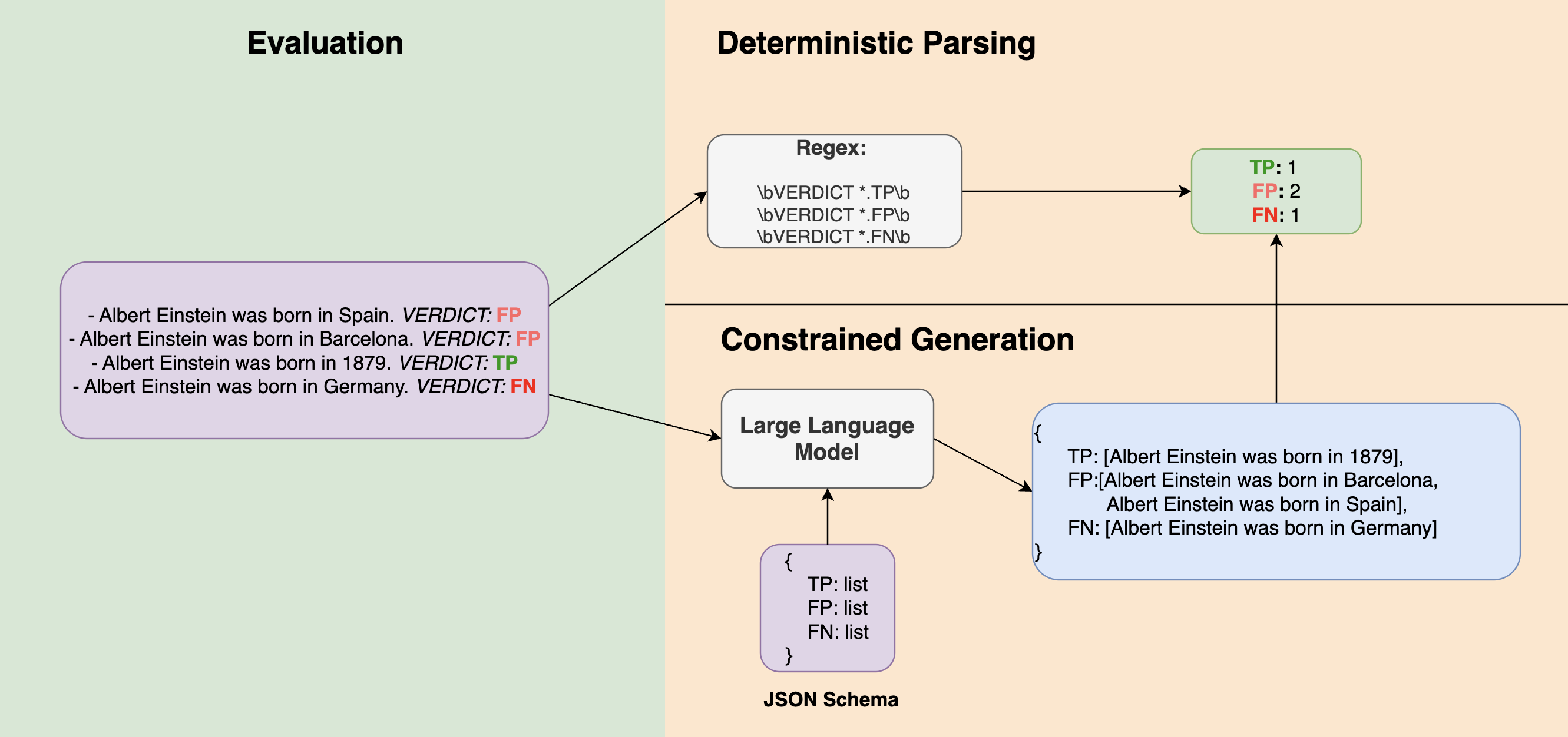}  
  \caption{The parsing strategies for extracting the labelled statements. The deterministic parsing uses regular expressions to match the labels. The constrained generation parsing collects the labelled statements in a JSON schema and then returns the number of matches for each label.}
  \label{fig:parsing_strategy}
\end{figure*}

\par The first regex (\textbf{Regex 1}) is trying to match the VERDICT keyword as well as the corresponding label. Therefore, depending on the calculated metric, we apply as many regular expressions as labels that need to be extracted.

For answer correctness:

\begin{itemize}
  \setlength{\itemsep}{2pt}
  \setlength{\parskip}{0pt}
  \item \textbf{True Positive}: \verb|\b|VERDICT: TP\verb|\b|
  \item \textbf{False Positive}: \verb|\b|VERDICT: FP\verb|\b|
   \item \textbf{False Negative}: \verb|\b|VERDICT: FN\verb|\b|
\end{itemize}

For faithfulness:

\begin{itemize}
  \setlength{\itemsep}{2pt}
  \setlength{\parskip}{0pt}
   \item \textbf{PASSED}: \verb|\b|VERDICT: PASSED\verb|\b|
   \item \textbf{FAILED}: \verb|\b|VERDICT: FAILED\verb|\b|
\end{itemize}

\par The second regex (\textbf{Regex 2}) matches the same pattern as the first regex, with the exception that it allows the possibility of having additional characters between the VERDICT keyword and the corresponding label.

For answer correctness: 

\begin{itemize}
  \setlength{\itemsep}{2pt}
  \setlength{\parskip}{0pt}
  \item \textbf{True Positive}: \verb|\b|VERDICT: .*TP\verb|\b|
  \item \textbf{False Positive}: \verb|\b|VERDICT: .*FP\verb|\b|
   \item \textbf{False Negative}: \verb|\b|VERDICT: .*FN\verb|\b|
\end{itemize}

For faithfulness: 

\begin{itemize}
  \setlength{\itemsep}{2pt}
  \setlength{\parskip}{0pt}
   \item \textbf{PASSED}: \verb|\b|VERDICT: .*PASSED\verb|\b|
   \item \textbf{FAILED}: \verb|\b|VERDICT: .*FAILED\verb|\b|
\end{itemize}

\par We propose the second version as a dynamic way of handling the scenarios when the LLM evaluator does not respect the first regex pattern.

\subsection{JSON Schemas}
\label{appendix:json}

\par Outlines requires a JSON schema from which a finite automata will be created. At each step during generation, the JSON schema will constrain the logits of the LLM evaluator by masking the invalid tokens. The process continues until the generation reaches an end state. Figure \ref{fig:outlines_simplified} shows an example of the final output.

\par The JSON schema for correctness is:

\begin{lstlisting}
{
    TP : list 
    FP : list 
    FN : list
}
\end{lstlisting}

\par The JSON schema for faithfulness is: 

\begin{lstlisting}
{
    PASSED : list 
    FAILED : list 
}
\end{lstlisting}

\subsection{Simplified Constrained Generation}
\label{sec:appendix_constrained}

\begin{figure}[h]  
  \centering
  \includegraphics[width=\linewidth]{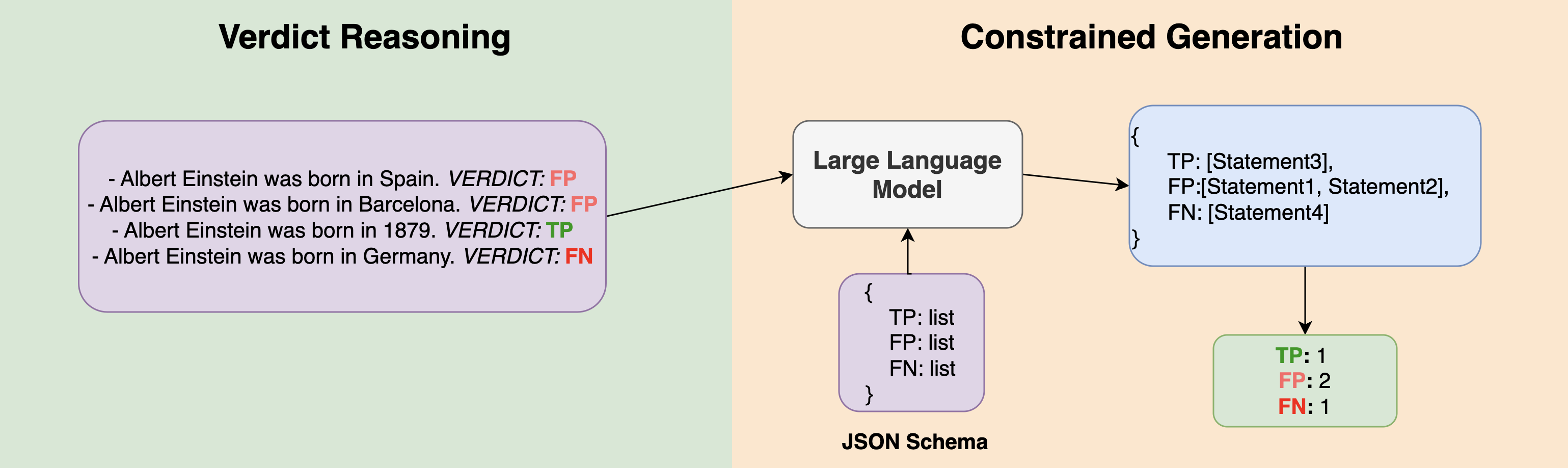}  
  \caption{Outlines simplified generation. Rather than copying each token of the statement in the list, the generation process is simplified by replacing the statement with a single token.}
  \label{fig:outlines_simplified}
\end{figure}

Figure \ref{fig:outlines_simplified} displays a simplified constrained parsing. Given few-shot examples and a JSON schema, the LLM evaluator maps the labelled statements to a single token and appends them to the corresponding list. 

\vspace{10mm}

\section{Prompts}
\label{appendix:prompts}

\subsection{Statement Extraction}
\vspace{5mm}
\begin{promptbox}
 Given a question, an answer, and sentences from the answer analyze the complexity of each sentence given under 'sentences' and break down each sentence into one or more fully understandable statements while also ensuring no pronouns are used in each statement. Format the output of the statements as a list with hyphens

  Examples: [\\
        \{\\
            "question": \\ "Who was Albert Einstein and what is he best known for?", \\
            "answer": "He was a German-born theoretical physicist, widely acknowledged to be one of the greatest and most influential physicists of all time. He was best known for developing the theory of relativity, he also made important contributions to the development of the theory of quantum mechanics.", \\
            "sentences":\\ "        
            0:He was a German-born theoretical physicist, widely acknowledged to be one of the greatest and most influential physicists of all time. \\     
            1:He was best known for developing the theory of relativity, he also made important contributions to the development of the theory of quantum mechanics. ", \\
            "statements": \\"                    
            - Albert Einstein was a German-born theoretical physicist. \\  - Albert Einstein is recognized as one of the greatest and most influential physicists of all time.   \\  
            - Albert Einstein was best known for developing the theory of relativity.    \\
            - Albert Einstein also made important contributions to the development of the theory of quantum mechanics.    "\\
        \}\\
    ],

Question: \{question\}

Answer: \{answer\}

Sentences: \{sentences\}
\\

 You need to output only the list of statements with hyphens. There can also be answers that contain only 1 statement. If you cannot find more than 1 statement, then just print the original answer with a hyphen.

\end{promptbox}

\subsection{Correctness Verdict}

\vspace{5mm}
\begin{promptbox}

Given a set of ground truth statements and a set of answer statements, analyze each statement and classify them in one of the following categories:

- TP (true positive): statements that are present in answer that are also supported by one or more statements in ground truth,

- FP (false positive): statements that are present in answer and that are not supported by any statements in ground truth,

- FN (false negative): statements that are present in the ground truth, but they are not supporting any statements in the answer.

Each statement can only belong to one of the categories. TP and FP are directly related to answer statements and FN directly related to ground truth statements. If a ground truth statement is supporting an answer statement, that statement can NEVER be an FN so avoid classifying it.

Examples: 
        [

        \{\\
            "question": "What powers the sun and what is its primary function?", \\
            "statements answers": \\" 
            - The sun is powered by nuclear fission, similar to nuclear reactors on Earth.   \\            
            - The primary function of the sun is to provide light to the solar system. ", \\
            "ground\_truth":\\ " 
            - The sun is powered by nuclear fusion, where hydrogen atoms fuse to form helium.\\               
            - This fusion process in the sun's core releases a tremendous amount of energy. \\             
            - The energy from the sun provides heat and light, which are essential for life on Earth.\\
            - The sun's light plays a critical role in Earth's climate system.  \\              
            - Sunlight helps to drive the weather and ocean currents.       ",

            "classification": \\"                     
            - The primary function of the sun is to provide light to the solar system. This statement is somewhat supported by the ground truth mentioning the sun providing light and its roles, though it focuses more broadly on the sun's energy. VERDICT: TP,\\              
            - The sun is powered by nuclear fission, similar to nuclear reactors on Earth. This statement is incorrect and contradicts the ground truth which states that the sun is powered by nuclear fusion. VERDICT: FP,  \\          
            - The sun is powered by nuclear fusion, where hydrogen atoms fuse to form helium. This accurate description of the sun’s power source is not included in the answer. VERDICT: FN    \\         
            - This fusion process in the sun's core releases a tremendous amount of energy. This process and its significance are not mentioned in the answer. VERDICT: FN \\
            - The energy from the sun provides heat and light, which are essential for life on Earth. The answer only mentions light, omitting the essential aspects of heat and its necessity for life, which the ground truth covers. VERDICT: FN      \\     
            - The sun's light plays a critical role in Earth's climate system. This broader impact of the sun’s light on Earth's climate system is not addressed in the answer. VERDICT: FN \\             
            - Sunlight helps to drive the weather and ocean currents. The effect of sunlight on weather patterns and ocean currents is omitted in the answer. VERDICT: FN" \\
        \}, 

        \{
            "question": "What is the boiling point of water?",\\
            "statements answers": \\ "      
            - The boiling point of water is 100 degrees Celsius at sea level   ", \\
            "statements ground\_truth": \\"    
            - The boiling point of water is 100 degrees Celsius (212 degrees Fahrenheit) at sea level.  \\    
            - The boiling point of water can change with altitude.", \\
            "classification": \\"          
            - The boiling point of water is 100 degrees Celsius at sea level. This statement is directly supported by the ground truth which specifies the boiling point of water as 100 degrees Celsius at sea level. VERDICT: TP,  \\
            - The boiling point of water can change with altitude. This additional information about how the boiling point of water can vary with altitude is not mentioned in the answer. VERDICT: FN           \\
            - The boiling point of water is 100 degrees Celsius (212 degrees Fahrenheit) at sea level. No need to label it because it is a ground truth statement that supports a statement from the answer.      " \\
        \},
        
        \{ \\
            "question": "Which actor is playing Han Solo in the original Star Wars?", \\
            "statements answers": \\"   
            - Han Solo is played by the American actor Harrison Ford.",\\
            "statements ground\_truth": \\
            "         
            - Harrison Ford    ", \\
            "classification": \\"               
            - Han Solo is played by the American actor Harrison Ford. This statement is directly supported by the ground which mentions Harrison Ford. VERDICT: TP  \\             - Harrison Ford. No need to label it because it is a ground truth statement that supports a statement from the answer.\\         "
        \}
        
        ]

Given the question and the statements, evaluate the input from below:

Question: \{question\}

Statements answer: \{statements\_answer\}

Statements ground\_truth: \{statements\_groundtruth\}

FP statements can be found only in the answer and FN statements can be found only in the ground truth. Remeber that if a statement is not present in the ground truth, then that is an FP not an FN!  If a ground truth statement is supporting an answer statement, that ground truth statement can NEVER be an FN so you no longer need to classify it. You don't need to look for the exact formulation of a ground truth statement inside an answer statement. Ground truth statements can be present inside an answer statement, but formulated in a different way and using different words (follow the examples I showed you above). Respect the structure of how you can give a verdict: [VERDICT: TP/FP/FN] 
\end{promptbox}

\subsection{Faithfulness Verdict}
\vspace{5mm}

\begin{promptbox}

Your task is to judge the faithfulness of a series of statements based on a given context. For each statement you must return verdict as PASSED if the statement can be directly inferred based on the context or FAILED if the statement can not be directly inferred based on the context.

Examples:  [ \\
        \{\\
            "context": \\
            "John is a student at XYZ University. He is pursuing a degree in Computer Science. He is enrolled in several courses this semester, including Data Structures, Algorithms, and Database Management. John is a diligent student and spends a significant amount of time studying and completing assignments. He often stays late in the library to work on his projects.", \\
            "statements": \\ "
            - John is majoring in Biology.,   \\                    
            - John is taking a course on Artificial Intelligence.,  \\ 
            - John is a dedicated student.,                     \\
            - John has a part-time job.", \\
            "answer": \\ "              
            - John is majoring in Biology. John's major is explicitly mentioned as Computer Science. There is no information suggesting he is majoring in Biology. VERDICT: FAILED \\
            - John is taking a course on Artificial Intelligence.The context mentions the courses John is currently enrolled in, and Artificial Intelligence is not mentioned. Therefore, it cannot be deduced that John is taking a course on AI. VERDICT: FAILED\"   \\ 
            - John is a dedicated student. The context states that he spends a significant amount of time studying and completing assignments. Additionally, it mentions that he often stays late in the library to work on his projects, which implies dedication. VERDICT: PASSED \\
            - John has a part-time job. There is no information given in the context about John having a part-time job. VERDICT: FAILED " \\
        \}, \\
        \{\\
            "context": "Photosynthesis is a process used by plants, algae, and certain bacteria to convert light energy into chemical energy.", \\
            "statements": \\
            "-Albert Einstein was a genius.", \\
            "answer": \\"                                            - Albert Einstein was a genius. The context and statement are unrelated. VERDICT: FAILED  " \\
        \}

Context: \{context\}\\
Statements: \{statements\}
  
  Include PASSED or FAILED only when you label a statement. Do not count, just label the statements. If something is not mentioned in the provided context, then it should be marked with FAILED. Every statements from the list provided needs to be evaluated. Respect the structure of how you can give a verdict to a statement: [VERDICT: PASSED/FAILED]",
\end{promptbox}

\subsection{Constrained Parsing}

\subsubsection{Correctness}

\vspace{5mm}

\begin{promptbox}
Given a set of statments that were labeled with either TP, FP or FN,  append the statement to the correct list. If a statement was labeled with TP tag, the statement should be appended to the TP list. If it was labeled with the FP tag, the statement should be in the FP list. If it was labeled with the FN tag, the statement should be in the FN list. I will provide you an example.

Example: 

 [\\
        \{\\
            "classified\_statements": \\ "    
            - Statement1 VERDICT: TP   \\
            - Statement2 VERDICT: TP   \\
            - Statement3 VERDICT: TP  \\
            - Statement4 VERDICT: FN \\
            - Statement5 VERDICT: FP \\
            - Statement6 VERDICT: FN  \\
            ",\\
            "output": \\"            
            \{               
            TP=[Statement1, Statement2, Statement3],   \\            
            FP=[Statement5],                \\
            FN=[Statement4, Statement6]                      
            \}       "\\
        \}, \\
        
        \{
            "classified\_statements": \\ "           
            - Statement1 VERDICT: FP     \\       
            - Statement2 VERDICT: TP       \\ 
            - Statement3 VERDICT: FN         \\  
            ",
            "output": \\"                 
            \{        
            TP=[Statement2],        \\
            FP=[Statement1],          \\   
            FN=[Statement3]          
            \}      
            "
        \}, \\
        
        \{
            "classified\_statements": \\"          
            - Statement1 VERDICT: FN         \\
            - Statement2 VERDICT: FP         \\
            - Statement3 VERDICT: TP   \\
            ",
            "output": \\ "                  
            \{              
            TP=[Statement3],  \\         
            FP=[Statement2],   \\    
            FN=[Statement1]   \\      
            \       
            "
        \}
    ]

Statements: \{statements\}

 Your task is only to put the number of the statements in the correct list. Sometimes the input you will receive will not be the same like the example, but try as best as you can to fulfill the task correctly.",

\end{promptbox}

\subsubsection{Faithfulness}
\vspace{5mm}

\begin{promptbox}
 Given a set of statments that were labeled with either PASSED OR FAILED,  append the statement to the correct list. If a statement was labeled with PASSED tag, the statement should be appended to the PASSED list. If it was labeled with the FAILED tag, the statement should be in the FAILED list. I will provide you an example.

Examples:

Example 1: \\
- Statement1 VERDICT: FAILED   \\ 
- Statement2 VERDICT: PASSED   \\ 
- Statement3 VERDICT: PASSED   \\ 
- Statement4 VERDICT: FAILED   \\             
Output:       \\  \{      
PASSED: [Statement2, Statement3]     \\
FAILED: [Statement1, Statement4]   
\}      \\

Example 2:      \\
- Statement1 VERDICT: PASSED   \\
- Statement2 VERDICT: PASSED    \\
- Statement3 VERDICT: FAILED   \\
- Statement4 VERDICT: PASSED   \\
- Statement5 VERDICT: FAILED  \\
Output:    \{      \\
PASSED: [Statement1, Statement2, Statement4] \\    
FAILED: [Statement3, Statement5] \\
\}        "

Statements : \{statements\}

Your task is only to put the number of the statements in the correct list. Sometimes the input you will receive will not respect strictly the example, but try as best as you can to fulfill the task correctly.,

\end{promptbox}

\end{document}